# On State Estimation in Multi-Sensor Fusion Navigation: Optimization and Filtering

Feng Zhu, Zhuo Xu, Xveqing Zhang, Yuantai Zhang, Weijie Chen, Xiaohong Zhang

**Abstract**—The essential of navigation, perception, and decision-making which are basic tasks for intelligent robots, is to estimate necessary system states. Among them, navigation is fundamental for other upper applications, providing precise position and orientation, by integrating measurements from multiple sensors. With observations of each sensor appropriately modelled, multi-sensor fusion tasks for navigation are reduced to the state estimation problem which can be solved by two approaches: optimization and filtering. Recent research has shown that optimization-based frameworks outperform filtering-based ones in terms of accuracy. However, both methods are based on maximum likelihood estimation (MLE) and should be theoretically equivalent with the same linearization points, observation model, measurements, and Gaussian noise assumption. In this paper, we deeply dig into the theories and existing strategies utilized in both optimization-based and filtering-based approaches. It is demonstrated that the two methods are equal theoretically, but this equivalence corrupts due to different strategies applied in real-time operation. By adjusting existing strategies of the filtering-based approaches, the Monte-Carlo simulation and vehicular ablation experiments based on visual odometry (VO) indicate that the strategy adjusted filtering strictly equals to optimization. Therefore, future research on sensor-fusion problems should concentrate on their own algorithms and strategies rather than state estimation approaches.

## 1 INTRODUCTION

Multi-sensor fusion for navigation, perception, and decision-making is a crucial prerequisite for autonomous systems, such as unmanned aerial vehicles (UAVs), autonomous cars, mobile robots, and other emerging applications [1–3], where standalone sensor is insufficient. All these fusion tasks are to extract relevant information from substantial volume of noisy measurements, using them to estimate necessary parameters of the system. Among these, positioning and navigation is a fundamental and indispensable module for other upper-layer applications [4–6]. Wrongly estimated position and orientation of autonomous robots will lead to life and property safety incidents. When measurements from each sensor are extracted and modelled appropriately, the positioning task is boiled down to state estimation problem which can be solved by optimization or filtering.

Essentially, the nature of optimization is least squares (LSQ), which has been widely utilized in solving precise positioning issues using global navigation satellite system (GNSS). The task of GNSS is to estimate position of antennas by signals from satellites, where the orbits and satellite clocks, are firstly generated by LSQ [7–9]. Then, using LSQ, user can achieve precise positioning results with observations appropriately modelled. Absolute position of stationary GNSS sites, or relative baseline between two arbitrary ones can also be precisely estimated by LSQ. For example, the world-famous scientific GNSS data processing software Bernese, developed by University of Bern [10], and GAMIT, developed by MIT [11], are based on LSQ, which are often used in geodesy.

However, due to vulnerable feature in urban viaducts and other signal blocked areas, the performance of GNSS degrades rapidly, leading to unreliable positioning results [12]. Fortunately, inertial measurement units (IMUs), which are independent of external infrastructure, are integrated with GNSS by extended Kalman filter (EKF), where IMU provides rigorous dynamic model by observing motion of carriers with raw measurements or absolute position of GNSS updating system states [13–16]. For example, Inertial Explorer (IE) and POSPac, which are excellent commercial software used to provide ground truth in navigation research, are developed based on EKF.

Meanwhile, EKF is also applied in simultaneous localization and mapping (SLAM) at an early stage in computer vision research. The first real-time SLAM system, MonoSLAM, is developed based on EKF [17]. Similarly, IMU is incorporated to recover scale and enable large-scale operation. One direct way to integrate measurements from cameras is to jointly estimate current IMU pose as well as visual landmarks, which results in high computational complexity [18,19]. The other way, a more time efficient and profoundly influential framework, adopts a sliding window filtering (SWF) to constrain multiple cameras and project visual bearing measurements onto the null space of the feature Jacobian matrix, rather than estimate them directly, namely MSCKF [20,21]. Based on MSCKF, researchers further utilize mapped landmarks as well as tracked ones in spacecraft landing and descent [22], to avoid wrong observability property for fast UAVs [23,24]. Besides, other filtering methods, such as unscented Kalman filter (UKF), particle filter (PF) et al., are employed to handle the high degree of nonlinearity and non-Gaussian noise [25–29]. In

- *Feng Zhu, Zhuo Xu, Xveqing Zhang, Yuantai Zhang, Weijie Chen, Xiaohong Zhang are with the School of Geodesy and Geomatics, Hubei Luojia Laboratory, Wuhan University, 129 Luoyu Road, Wuhan 430079, Hubei, China. E-mail: {fzhu, zhuoxu, xveqing, officialtai, wjchen}@whu.edu.cn, xhzhang@sgg.whu.edu.cn.*
- *Feng Zhu is also with the Key Laboratory of Geospace Environment and Geodesy, Ministry of Education, Wuhan University, 129 Luoyu Road, Wuhan 430079, China.*
- *Xiaohong Zhang is also with Chinese Antarctic Center of Surveying and Mapping, Wuhan University, 129 Luoyu Road, Wuhan 430079, Hubei, China.*

*This study was supported by the National Natural Science Foundation of China (Grant No. 42374031), the National Natural Science Foundation of China (Grant No. 42104021), and the Special Fund of Hubei Luojia Laboratory (Grant No. 2201000038).*
*(Corresponding author: Feng Zhu)*

      

[30,31], invariant EKF (InEKF) has been introduced to preserve the observable structure of the navigation system. Nevertheless, most matured and successful approaches in filtering have assumed Gaussian noise [32], making MSCKF still the mainstream of filtering-based approaches.

On the other hand, optimization-based approaches have flourished in recent research of computer vision, as there are open-source libraries, such as Ceres, g²o, GTSAM et al., offering abilities of plugin-and-use, outlier removal, and acceleration, where the estimating states and measurements are respectively abstracted into vertices and edges by a graph to intuitively express the state estimation problem [33–36]. Even numerous advantages are featured, the foundation of optimization-based methods still fall under the LSQ. Additionally, although incremental approaches, such as incremental smoothing [34], are applied to reuse previous calculations to improve efficiency, they are still suffered from increasing size of the graph, leading to computational burdens. On that account, sliding window optimization (SWO) is widely adopted to achieve constant processing time as well as to make full use of measurements in SLAM area [5,37–40]. In order to incorporate historic information in the sliding window, SWO utilizes marginalization when window slides, i.e. Schur complement, where information of removed nodes is transferred to its Markov blanket as prior information [41]. Similar to filtering-based approaches, graph of incorporating absolute information from GNSS is constructed, including both state space representation [42,43], and observation space representation [44–46], to mitigate drifts of visual and inertial navigation systems based on SWO.

Comparing research in GNSS positioning with computer vision, it is surprising to note that optimization is preferred in computer vision while filtering is popular in positioning. [32] proposes a combined accuracy indicator to compare EKF with optimization and its experimental results show that optimization outperforms filtering based on SLAM. Other frameworks based on optimization also shows consistent conclusions that optimization performs better compared with filtering, but detailed analysis is not presented [38,43,44,47,48].

However, both filtering and optimization can be derived from maximum likelihood estimation (MLE) by assuming Gaussian noise [49,50], which indicates that optimization-based and filtering-based approaches should be mathematically equivalent under the assumption of the same measurements and observation model. Even though real-time frameworks utilize sliding window to bound computational complexity, the state estimators are still LSQ or EKF. Besides, null-space marginalization is applied in MSCKF, while Schur complement is used in SWO, the equivalence between them has been analytically demonstrated in [51]. Obviously, it is contradictory that both theories and tricks utilized in the two approaches are equivalent, yet not the estimates.

In order to find out reasons causing these discrepancies and realize potential estimating accuracy in the filtering, the two state estimation methods are theoretically and comprehensively analyzed in the paper, including state estimation theory and strategies applied in sliding window estimation, such as marginalization and first estimate Jacobian (FEJ). Then, based on the detailed analysis, existing strategies utilized in filtering-based methods are adjusted by a 2-step update strategy, which shows the same performance with optimization-based ones. To evaluate the performance of strategy adjusted filtering, a series of Monte-Carlo simulation and vehicular ablation experiments are conducted where data collected by KITTI [52]. Specifically, the main contributions of this work include:

1) Discrepancies between optimization and filtering are theoretically analyzed in depth, including prior constraints, measurement utilization, linearization points, and common strategies utilized in real-time estimation, such as marginalization and FEJ. This analysis mathematically demonstrates that the differences are caused by strategies, which further leads to different constraints and measurement utilization, rather than state estimation methods.

2) Based on theoretical analysis, existing strategies in the filtering are adjusted by a 2-step update strategy, which fully incorporates constraints from previous window, and FEJ compensation.

3) Extensive experiments are performed on both Monte-Carlo simulation and ablation experiments using KITTI dataset based on visual odometry. The results show that the 2-step update strategy makes filtering equal to optimization under the mild assumption of independent and identically distributed (i.i.d.) Gaussian noise and the same measurements utilization when solving the same state estimation problem.

The rest of this paper is structured as follows. In section 2, notations and two alternative representations of the Kalman filter are introduced. In section 3, equivalence and potential discrepancies are discussed between batch estimation and sequential update. Section 4 further analyzes prior constraints, measurement utilization and FEJ compensation in sliding window cases. In section 5, a 2-step modification is proposed to adjust existing strategies in sliding window filtering based on previous analysis. In section 6, both simulation and ablation experiments are performed to validate the preceding analysis and the performance of strategy adjusted filtering-based method. Finally, this paper is concluded in section 7.

## 2 PRELIMINARIES

In this section, notations used throughout this paper is briefly introduced. And then two representations of the Kalman filter (KF) are mathematically derived, to simplify the subsequent analysis.

In order to analyze state estimation of the optimization-based and filtering-based approaches, assume $X = \begin{bmatrix} x_1^T & \cdots & x_M^T \end{bmatrix}^T$ is going to be estimated with $M$ timestamps involved. Specifically, $\hat{x}_t$ denotes the estimate of a variable $x_t$ at timestamp $t$ and $\tilde{x}_t$ denotes an observation of $x_t$. The error state $\delta x_t$ of $x_t$ is always estimated throughout this paper, which is defined by

$$\delta x_t = \tilde{x}_t - x_t. \tag{1}$$

$X$ can be estimated only when measurements are all available, where $Z = \{z_1, z_2, \ldots, z_M\}$ denotes measurement set with $M$ timestamps included. In the sliding window cases, $X_t = \begin{bmatrix} x_t^T & \cdots & x_{t+N}^T \end{bmatrix}^T$ and $Z_t = \{z_t, z_{t+1}, \ldots, z_{t+N}\}$ denotes $N$ involved states and measurements at timestamp $t$ respectively.

The size of $\mathbf{Z}$ and $\mathbf{Z}_t$ can be expressed by $s$ and $s_t$. For each measurement $\mathbf{z}_t$, it can be modelled as

$$\mathbf{z}_t = \mathbf{f}_t(\mathbf{x}_t) + \mathbf{n}_t. \tag{2}$$

Where, $\mathbf{f}_t(\cdot)$ is the measurement function, mapping from system states to measurements. And $\mathbf{n}_t$ denotes the noise vector, modelled as zero-mean Gaussian noise with $\mathbf{n}_t \sim \mathcal{N}(\mathbf{0}, \mathbf{\Lambda}_t^{-1})$. $\mathbf{\Lambda}_t$ denotes the measurement weight matrix. By employing Taylor-series approximation, this observation model can be approximated as (3) with error state expression.

$$\delta\mathbf{z}_t = \mathbf{h}_t \delta\mathbf{x}_t + \mathbf{o}(\mathbf{x}_t) + \mathbf{n}_t. \tag{3}$$

Where, $\mathbf{o}(\mathbf{x}_t)$ denotes the higher order terms, and $\mathbf{h}_t$ denotes the Jacobian matrix. In the linear case, there only exists first order terms, and $\mathbf{o}(\mathbf{x}_t)$ always equals to $\mathbf{0}$. While, for the nonlinear system, $\mathbf{o}(\mathbf{x}_t)$ could be ignored during state estimation.

Additionally, system states can be predicted by dynamic model, which can be expressed by

$$\mathbf{x}_{t|t-1} = \mathbf{g}_t(\mathbf{x}_{t-1}) + \mathbf{w}_t. \tag{4}$$

Similar with (3), by employing Taylor-series approximation and discretization, the linearized and time-discrete dynamic model can be approximated as (5) with error state expression.

$$\delta\mathbf{x}_{t|t-1} = \mathbf{\Phi}_t \delta\mathbf{x}_{t-1} + \mathbf{w}_t. \tag{5}$$

Where, $\mathbf{w}_t$ denotes a noise vector with covariance matrix $\mathbf{Q}_t$, which is also modelled as Gaussian noise $\mathbf{w}_t \sim \mathcal{N}(\mathbf{0}, \mathbf{Q}_t)$. Suppose $\mathbf{P}_{t-1}$ describes the covariance of $\delta\mathbf{x}_{t-1}$, and its inverse $\mathbf{N}_{t-1}$ denotes the information matrix. The uncertainty of $\delta\mathbf{x}_{t|t-1}$ can be derived by:

$$\mathbf{P}_{t|t-1} = \mathbf{\Phi}_t \mathbf{P}_{t-1} \mathbf{\Phi}_t^T + \mathbf{Q}_t. \tag{6}$$

By regarding $\delta\mathbf{x}_{t|t-1}$ and $\mathbf{P}_{t|t-1}$ as a virtual measurement, $\delta\mathbf{x}_t$ can be estimated through MLE:

$$\delta\hat{\mathbf{x}}_t = \arg\max p(\delta\mathbf{z}_{t-1,p}, \delta\mathbf{z}_t | \delta\mathbf{x}_t). \tag{7}$$

Where, the virtual measurement $\delta\mathbf{z}_{t-1,p}$ has the following form with $\mathbf{w}_{t-1,p} \sim \mathcal{N}(\mathbf{0}, \mathbf{P}_{t|t-1})$.

$$\delta\mathbf{z}_{t-1,p} = \delta\mathbf{x}_{t|t-1} + \mathbf{w}_{t-1,p}. \tag{8}$$

Under Gaussian noise assumption, this MLE can be solved by LSQ, where the principal is to minimize

$$\begin{bmatrix} \delta\mathbf{z}_{t-1,p} \\ \delta\mathbf{z}_t \end{bmatrix}^T \begin{bmatrix} \mathbf{P}_{t|t-1}^{-1} & \\ & \mathbf{\Lambda}_t \end{bmatrix} \begin{bmatrix} \delta\mathbf{z}_{t-1,p} \\ \delta\mathbf{z}_t \end{bmatrix} = \min. \tag{9}$$

Considering (3) and (8), LSQ (10) can be further derived from (9), which also denotes the information filter.

$$\delta\hat{\mathbf{x}}_t = \left(\mathbf{P}_{t|t-1}^{-1} + \mathbf{h}_t^T \mathbf{\Lambda}_t \mathbf{h}_t\right)^{-1} \left(\mathbf{P}_{t|t-1}^{-1} \delta\mathbf{x}_{t|t-1} + \mathbf{h}_t^T \mathbf{\Lambda}_t \mathbf{l}_t\right). \tag{10}$$

Where, $\mathbf{l}_t$ denotes the residual related part, which has the following form

$$\mathbf{l}_t = \mathbf{z}_t - \mathbf{f}\left(\hat{\mathbf{x}}_{t|t-1}\right),$$
$$\mathbf{b}_t = \mathbf{P}_{t|t-1}^{-1} \delta\mathbf{x}_{t|t-1} + \mathbf{h}_t^T \mathbf{\Lambda}_t \mathbf{l}_t. \tag{11}$$

And, $\hat{\mathbf{x}}_{t|t-1} = \mathbf{g}_t(\hat{\mathbf{x}}_{t-1})$ describes the propagated system states.

Additionally, the posteriori of the LSQ can be derived by

$$\mathbf{N}_t = \mathbf{P}_{t|t-1}^{-1} + \mathbf{h}_t^T \mathbf{\Lambda}_t \mathbf{h}_t = \mathbf{P}_t^{-1}. \tag{12}$$

Then, (10) can be rewritten as

$$\delta\mathbf{x}_t = \mathbf{P}_t \left(\left(\mathbf{P}_t^{-1} - \mathbf{h}_t^T \mathbf{\Lambda}_t \mathbf{h}_t\right) \delta\mathbf{x}_{t|t-1} + \mathbf{h}_t^T \mathbf{\Lambda}_t \mathbf{l}_t\right). \tag{13}$$

Thus, we have

$$\delta\hat{\mathbf{x}}_t = \delta\mathbf{x}_{t|t-1} + \mathbf{P}_t \mathbf{h}_t^T \mathbf{\Lambda}_t \left(\mathbf{l}_t - \mathbf{h}_t \delta\mathbf{x}_{t|t-1}\right). \tag{14}$$

Considering

$$\left(\mathbf{D}^{-1} + \mathbf{C}^T \mathbf{A} \mathbf{C}\right)^{-1} \mathbf{C}^T \mathbf{A} = \mathbf{D} \mathbf{C}^T \left(\mathbf{A}^{-1} + \mathbf{C} \mathbf{D}^{-1} \mathbf{C}^T\right)^{-1}, \tag{15}$$

coefficient in (14) can be further denoted as

$$\mathbf{P}_t \mathbf{h}_t^T \mathbf{\Lambda}_t = \left(\mathbf{P}_{t|t-1}^{-1} + \mathbf{h}_t^T \mathbf{\Lambda}_t \mathbf{h}_t\right)^{-1} \mathbf{h}_t^T \mathbf{\Lambda}_t$$
$$= \mathbf{P}_{t|t-1} \mathbf{h}_t^T \left(\mathbf{\Lambda}_t^{-1} + \mathbf{h}_t \mathbf{P}_{t|t-1} \mathbf{h}_t^T\right)^{-1} \triangleq \mathbf{K}_t. \tag{16}$$

And the posteriori covariance matrix can be rewritten as:

$$\mathbf{P}_t = \left(\mathbf{P}_{t|t-1}^{-1} + \mathbf{h}_t^T \mathbf{\Lambda}_t \mathbf{h}_t\right)^{-1}$$
$$= \mathbf{P}_{t|t-1} - \mathbf{P}_{t|t-1} \mathbf{h}_t^T \left(\mathbf{\Lambda}_t^{-1} + \mathbf{h}_t \mathbf{P}_{t|t-1} \mathbf{h}_t^T\right) \mathbf{h}_t \mathbf{P}_{t|t-1}$$
$$= \left(\mathbf{I} - \mathbf{K}_t \mathbf{h}_t\right) \mathbf{P}_{t|t-1}. \tag{17}$$

(14), (16), and (17) form the formula of the KF, which is derived from (10). Thus, (10) and (14) are two alternative expressions of KF, namely information filtering representation (IFR) and Kalman filtering representation (KFR). Note that even the two expressions are equivalent, their time complexity is dependent on matrix inverse operation, which is approximately $O(n^3)$ in IFR, and $O(m^3)$ in KFR, where $n$ and $m$ are number of system states and measurements respectively.

Besides, KF only estimates states at a specific timestamp, but multiple variables from different timestamps can be jointly estimated in order to incorporate more information from measurements. Thus, batch estimation and sliding window strategy will be analyzed in the subsequent sections, which can be expressed with the two equivalent expressions as well. Then, (18) denotes the stacked Jacobi and residual vector in batch estimation and (19) denotes the counterparts in sliding window cases.

$$\mathbf{H} = \begin{bmatrix} \mathbf{h}_1^T & \cdots & \mathbf{h}_M^T \end{bmatrix}^T, \mathbf{L} = \begin{bmatrix} \mathbf{l}_1^T & \cdots & \mathbf{l}_M^T \end{bmatrix}^T, \tag{18}$$

$$\mathbf{H}_t = \begin{bmatrix} \mathbf{h}_t^T & \cdots & \mathbf{h}_{t+N}^T \end{bmatrix}^T, \mathbf{L}_t = \begin{bmatrix} \mathbf{l}_t^T & \cdots & \mathbf{l}_{t+N}^T \end{bmatrix}^T. \tag{19}$$

When a specific state estimation problem is expressed with either IFR or KFR, it is equivalent to the other representation. Thus, the two representations establish a bridge to analyze the equivalence between the optimization and filtering, which will be utilized to dig into the reasons that cause discrepancies between the optimization-based and filtering-based approaches in the subsequent analysis.

## 3 THEORETICAL ANALYSES ON OPTIMIZATION-BASED AND FILTERING-BASED APPROACHES

In this section, equivalence between batch estimation and se-quential update is firstly analyzed with a linear system, which ensures the same Jacobian matrix. Nonlinear system is then analyzed with iteration strategy applied.

### 3.1 Equivalence Between Optimization and Filtering in Linear System

Toward the state estimation problem with measurement set $\boldsymbol{Z}$ in a linear system, the optimal solution is to solve the full-state MLE problem, which jointly estimates all variables

$$\delta \hat{\boldsymbol{X}} = \arg\max p\left(\delta \boldsymbol{Z}_p, \delta \boldsymbol{Z} \mid \delta \boldsymbol{X}\right). \quad (20)$$

Similarly, regarding prior knowledge $\delta \boldsymbol{X}_0$ as a virtual measurement with noise vector $\boldsymbol{w}_p \sim \mathcal{N}\left(\boldsymbol{0}, \boldsymbol{P}_0\right)$, this MLE can be solved by full-state LSQ (21).

$$\delta \hat{\boldsymbol{X}} = \left(\boldsymbol{P}_0^{-1} + \boldsymbol{H}^T \boldsymbol{\Lambda} \boldsymbol{H}\right)^{-1}\left(\boldsymbol{P}_0^{-1} \delta \boldsymbol{X}_0 + \boldsymbol{H}^T \boldsymbol{\Lambda} \boldsymbol{L}\right). \quad (21)$$

Since all measurements are stacked to estimate states, it is also known as batch estimation. In practice, measurements are often time-irrelevant, such as GNSS pseudorange or phase measurements, visual feature points, and LiDAR scans et al. Thus, weight matrix $\boldsymbol{\Lambda}$ satisfies:

$$\boldsymbol{\Lambda} = diag\left\{\boldsymbol{\Lambda}_1 \quad \cdots \quad \boldsymbol{\Lambda}_M\right\}. \quad (22)$$

Then, considering (18), batch estimation can be rewritten as

$$\left(\boldsymbol{P}_0^{-1} + \sum_{t=1}^{M} \boldsymbol{h}_t^T \boldsymbol{\Lambda}_t \boldsymbol{h}_t\right)\delta \hat{\boldsymbol{X}} = \boldsymbol{P}_0^{-1}\delta \boldsymbol{X}_0 + \sum_{t=1}^{M} \boldsymbol{h}_t^T \boldsymbol{\Lambda}_t \boldsymbol{l}_t. \quad (23)$$

Besides, for each measurement in $\boldsymbol{Z}$, system states can be sequentially estimated by:

$$\begin{aligned}
\delta \hat{\boldsymbol{X}}_1 &= \left(\boldsymbol{P}_0^{-1} + \boldsymbol{h}_1^T \boldsymbol{\Lambda}_1 \boldsymbol{h}_1\right)^{-1}\left(\boldsymbol{P}_0^{-1}\delta \boldsymbol{X}_0 + \boldsymbol{h}_1^T \boldsymbol{\Lambda}_1 \boldsymbol{l}_1\right) \\
\delta \hat{\boldsymbol{X}}_2 &= \left(\boldsymbol{P}_1^{-1} + \boldsymbol{h}_2^T \boldsymbol{\Lambda}_2 \boldsymbol{h}_2\right)^{-1}\left(\boldsymbol{P}_1^{-1}\delta \hat{\boldsymbol{X}}_1 + \boldsymbol{h}_2^T \boldsymbol{\Lambda}_2 \boldsymbol{l}_2\right) \\
&\vdots \\
\delta \hat{\boldsymbol{X}}_M &= \left(\boldsymbol{P}_{M-1}^{-1} + \boldsymbol{h}_M^T \boldsymbol{\Lambda}_M \boldsymbol{h}_M\right)^{-1}\left(\boldsymbol{P}_{M-1}^{-1}\delta \hat{\boldsymbol{X}}_{M-1} + \boldsymbol{h}_M^T \boldsymbol{\Lambda}_M \boldsymbol{l}_M\right),
\end{aligned} \quad (24)$$

which is equivalent to the full-state linear KF (LKF) expressed by IFR. Considering the posteriori of the final estimates $\delta \hat{\boldsymbol{X}}_M$, we have

$$\begin{aligned}
\boldsymbol{P}_M^{-1} &= \boldsymbol{P}_{M-1}^{-1} + \boldsymbol{h}_M^T \boldsymbol{\Lambda}_M \boldsymbol{h}_M \\
&= \boldsymbol{P}_{M-2}^{-1} + \boldsymbol{h}_{M-1}^T \boldsymbol{\Lambda}_{M-1} \boldsymbol{h}_{M-1} + \boldsymbol{h}_M^T \boldsymbol{\Lambda}_M \boldsymbol{h}_M \\
&= \boldsymbol{P}_0^{-1} + \sum_{t=1}^{M} \boldsymbol{h}_t^T \boldsymbol{\Lambda}_t \boldsymbol{h}_t,
\end{aligned} \quad (25)$$

and the residual related part can be rewritten as

$$\begin{aligned}
\boldsymbol{b}_M &= \boldsymbol{P}_{M-1}^{-1}\delta \hat{\boldsymbol{X}}_{M-1} + \boldsymbol{h}_M^T \boldsymbol{\Lambda}_M \boldsymbol{l}_M \\
&= \boldsymbol{P}_{M-2}^{-1}\delta \hat{\boldsymbol{X}}_{M-2} + \boldsymbol{h}_{M-1}^T \boldsymbol{\Lambda}_{M-1} \boldsymbol{l}_{M-1} + \boldsymbol{h}_M^T \boldsymbol{\Lambda}_M \boldsymbol{l}_M \\
&= \boldsymbol{P}_0^{-1}\delta \boldsymbol{X}_0 + \sum_{t=1}^{M} \boldsymbol{h}_t^T \boldsymbol{\Lambda}_t \boldsymbol{l}_t.
\end{aligned} \quad (26)$$

Thus, final estimates in sequential update can be rewritten as (27), which shares the same mathematical form with the batch estimation.

$$\delta \hat{\boldsymbol{X}}_M = \left(\boldsymbol{P}_0^{-1} + \sum_{t=1}^{M} \boldsymbol{h}_t^T \boldsymbol{\Lambda}_t \boldsymbol{h}_t\right)^{-1}\left(\boldsymbol{P}_0^{-1}\delta \boldsymbol{X}_0 + \sum_{t=1}^{M} \boldsymbol{h}_t^T \boldsymbol{\Lambda}_t \boldsymbol{l}_t\right). \quad (27)$$

**Proposition 1**: The full-state MLE problem can be solved by batch estimation, or full-state LSQ (21), under Gaussian assumption. Supposing measurements are time-irrelevant, final estimates of sequential update, or full-state LKF (27), further equals to batch estimation (21) as well.

### 3.2 Jacobi Inconsistency in Nonlinear System

Due to linearization error introduced by Taylor-series approximation, an iterative approach should be applied. For $i$-th iteration at timestamp $t$, system states can be estimated by

$$\delta \hat{\boldsymbol{x}}_t^{(i)} = \left(\boldsymbol{P}_{t|t-1}^{-1} + \boldsymbol{h}_t^{(i)T} \boldsymbol{\Lambda}_t \boldsymbol{h}_t^{(i)}\right)^{-1}\left(\boldsymbol{P}_{t|t-1}^{-1}\delta \hat{\boldsymbol{x}}_t^{(i-1)} + \boldsymbol{h}_t^{(i)T} \boldsymbol{\Lambda}_t \boldsymbol{l}_t^{(i)}\right), \quad (28)$$

which denotes the nonlinear squares (NLSQ) and where $\delta \hat{\boldsymbol{x}}_t^{(0)} = \delta \boldsymbol{x}_{t|t-1}$ for the first iteration.

Meanwhile, the optimal solution of the nonlinear system can be obtained by jointly estimating all variables as well. The batch estimation solution of $i$-th iteration can be obtained by

$$\delta \hat{\boldsymbol{X}}^{(i)} = \left(\boldsymbol{P}_0^{-1} + \boldsymbol{H}^{(i)T} \boldsymbol{\Lambda} \boldsymbol{H}^{(i)}\right)^{-1}\left(\boldsymbol{P}_0^{-1}\delta \boldsymbol{X}^{(i-1)} + \boldsymbol{H}^{(i)T} \boldsymbol{\Lambda} \boldsymbol{L}^{(i)}\right). \quad (29)$$

Different from linear system, the Jacobian matrix is affected by system states, which is evaluated at the estimates of previous iteration:

$$\boldsymbol{H}^{(i)} = \left[\boldsymbol{h}_1^{(i)T}\Big|_{\boldsymbol{X}=\hat{\boldsymbol{X}}^{(i-1)}} \quad \cdots \quad \boldsymbol{h}_M^{(i)T}\Big|_{\boldsymbol{X}=\hat{\boldsymbol{X}}^{(i-1)}}\right]^T. \quad (30)$$

With i.i.d. Gaussian noise assumption, where (22) is satisfied, batch estimation in nonlinear system (29) can be rewritten as

$$\left(\boldsymbol{P}_0^{-1} + \sum_{t=1}^{M} \boldsymbol{h}_t^{(i)T} \boldsymbol{\Lambda}_t \boldsymbol{h}_t^{(i)}\right)\delta \hat{\boldsymbol{X}}^{(i)} = \boldsymbol{P}_0^{-1}\delta \boldsymbol{X}^{(i-1)} + \sum_{t=1}^{M} \boldsymbol{h}_t^{(i)T} \boldsymbol{\Lambda}_t \boldsymbol{l}_t^{(i)}. \quad (31)$$

Then by analogy with (24), the final estimates after M-th measurement using sequential update can be derived by

$$\delta \boldsymbol{X}_M^{(i)} = \left(\boldsymbol{P}_{M-1}^{-1} + \boldsymbol{h}_M^{(i)T} \boldsymbol{\Lambda}_M \boldsymbol{h}_M^{(i)}\right)^{-1}\left(\boldsymbol{P}_{M-1}^{-1}\delta \hat{\boldsymbol{X}}_{M-1}^{(i-1)} + \boldsymbol{h}_M^{(i)T} \boldsymbol{\Lambda}_M \boldsymbol{l}_M^{(i)}\right), \quad (32)$$

which also denotes the full-state iterative extended Kalman filter (IEKF) expressed by IFR and theoretically equals to the full-state NLSQ, i.e. batch estimation in nonlinear system (29). However, the sequential update iterates (32) until it converges at each measurement, and then compensates the estimates. Thus, Jacobi of next coming measurement is evaluated at previous estimates instead of initial value. For analysis, by stacking Jacobian matrices of each measurement update at i-th iteration in the sequential update, we have:

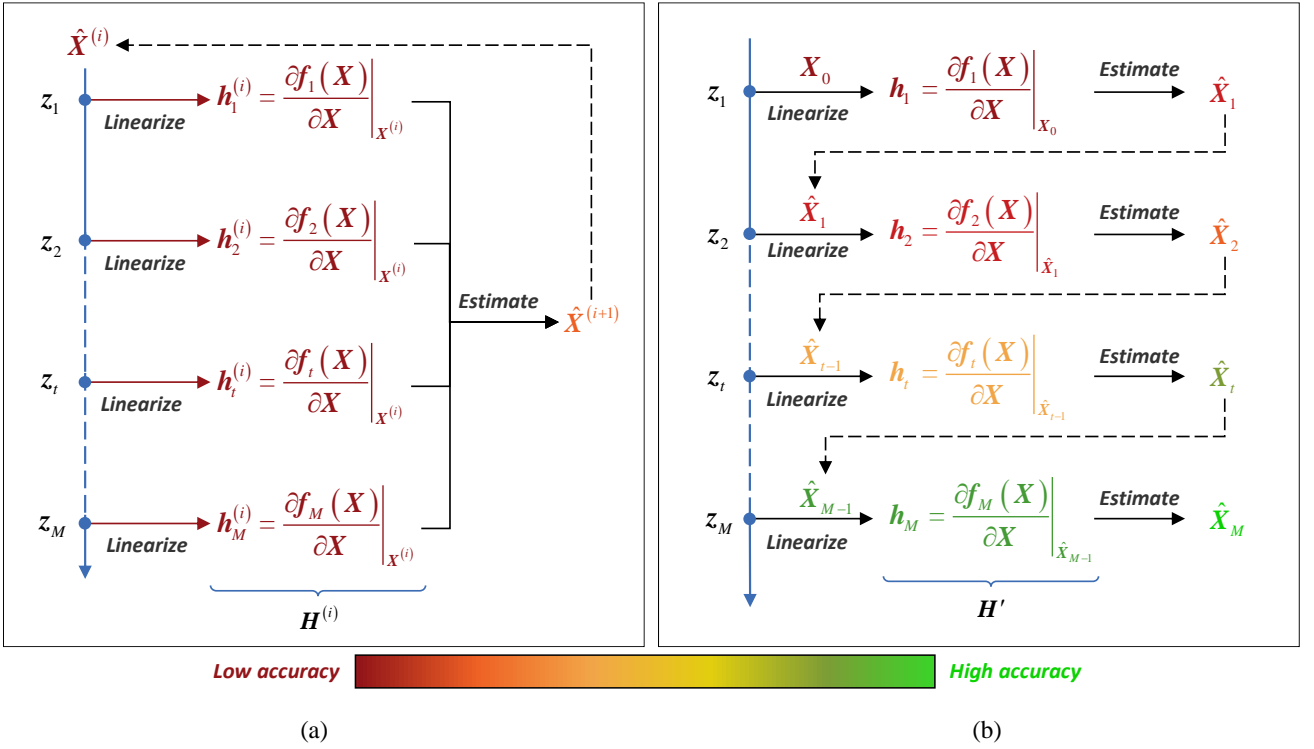

(a)                                                    (b)

**Fig. 1.** Illustration of the Jacobian inconsistency using batch estimation and sequential update in a nonlinear system. Batch estimation evaluates Jacobian matrix at the same linearization points, which is consistent (a). On the contrary, the sequential update iterates and compensates at each measurement, leading to different linearization points (b). Thus, the Jacobi of sequential update is inconsistent. Color represents the accuracy of linearization points and the estimates.

$$H'^{(i)} = \left[ h_1'^{(i)T} \big/_{X = \hat{X}_1^{(i-1)}} \quad \cdots \quad h_M'^{(i)T} \big/_{X = \hat{X}_M^{(i-1)}} \right]^T. \quad (33)$$

Where, $\hat{X}_t^{(i-1)}$ denotes the estimates using t-th measurement after iteration $i-1$. Compared (30) with (33), sequential update corrupts the consistency of the Jacobian matrix at the view of linearization points, as illustrated in Fig. 1. This inconsistency leads to wrong observations $\Delta H$ in the Jacobi and estimates $\Delta X$, as in (34).

$$Z = H^{(i)}\hat{X}^{(i)} = H'^{(i)}\hat{X}'^{(i)} = \left( H^{(i)} + \Delta H \right)\left( \hat{X}^{(i)} + \Delta X \right) \quad (34)$$

Thus, $\Delta X$ can be expressed by the inconsistency of the Jacobi, which implies that wrongly estimated states are severely affected by linearization points.

$$\Delta X = -\left( H^{(i)} + \Delta H \right)^{-g} \Delta H \hat{X}^{(i)} \quad (35)$$

Where, $-g$ is generalized inverse operation.

In order to achieve optimal estimates and avoid wrong observations introduced by inconsistency of the Jacobi. The feasible strategy is to update states measurement by measurement, which is regarded as once iteration (see Algorithm 1). Then, the Jacobi and residual vector are evaluated at compensated estimates in the next iteration, which ensures the same Jacobian matrix and residual vector with (33).

---

**Algorithm 1** Full-state Sequential Update Using KFR

set $i \leftarrow 0, t \leftarrow 1, MaxIter, \delta X_0^{(0)} \leftarrow \delta X^{(0)}, P_0$

**while** $i < MaxIter$

   **while** $t \leq M$

      // evaluate Jacobi and residual at $X^{(i)}$

$$h_t^{(i)} \leftarrow \frac{\partial f_t(X)}{\partial X}\bigg|_{X^{(i)}}, \ l_t^{(i)} \leftarrow z_t - f_t\left(X^{(i)}\right)$$

      // full-state sequential update

$$K_t^{(i)} \leftarrow P_{t-1}h_t^{(i)T}\left( \Lambda_t^{-1} + h_t^{(i)}P_{t-1}h_t^{(i)T} \right)^T \quad (36)$$

$$\delta X_t^{(i)} \leftarrow \delta X_{t-1}^{(i)} + K_t^{(i)}\left( l_t^{(i)} - h_t^{(i)}\delta X_{t-1}^{(i)} \right) \quad (37)$$

$$P_t \leftarrow \left( I - K_t^{(i)}h_t^{(i)} \right)P_{t-1} \quad (38)$$

   **end while**

$$\hat{X}_0^{(i+1)} \leftarrow X_0^{(i)} - \delta X_M^{(i)}$$

   **if** converge

      **return** $\hat{X}_0^{(i+1)}, P_M$

   **end if**

   $i \leftarrow i + 1$

**end while**

---

*Proposition 2*: For nonlinear system, NLSQ (29) and IEKF (32) are theoretically equivalent as well, although the Jacobian matrix is dependent on linearization points. Due to iteration strategies, estimating states using all measurements, compensating the estimates and then doing next iteration in sequential update (37) should be applied, which remains the consistency of the Jacobi compared with batch estimation. Instead, sequential update iterates and compensates at each measurement (32), which leads to Jacobian inconsistency (35).

Based on previous analysis, some conceptions can be clarified here. Batch estimation, i.e. full-state LSQ or NLSQ, falls under optimization, where they mathematically share the same formula but iteration strategy is applied in nonlinear cases. Meanwhile, sequential update, i.e. full-state LKF or IEKF, is a category of filtering. But note that the EKF is different from IEKF, where IEKF adopts iteration strategy while EKF not. Thus, IEKF equals to NLSQ but EKF not.

## 4 SLIDING WINDOW STRATEGY APPLIED IN OPTIMIZATION-BASED AND FILTERING-BASED APPROACHES

In this section, discrepancies caused by sliding window strategy, which is used to balance computational burdens with the accuracy, are discussed. Since sliding window optimization (SWO) utilizes information matrix while sliding window filtering (SWF) operates on covariance matrix in practice, IFR and KFR are utilized respectively in SWO and SWF for analysis.

First, estimates in the current window $t$ are analyzed. The MLE problem of the SWO or SWF at current window can be described by:

$$\delta \hat{X}_t = \arg \max p\left(\delta Z_{t,p}, \delta Z_t \middle| \delta \hat{X}_t\right). \tag{39}$$

Compared with (20), SWO or SWF is a case of batch estimation, where states in the window are jointly estimated. Therefore, parameters in the current window can be estimated equivalently, i.e. $\hat{X}_t$, provided the same initial values of parameters, measurements, observation model, and stochastic model in both estimation methods according to proposition 2.

After attaining the estimates, window will slide to maintain the same computational burdens. Since state estimation is equivalent in both SWO and SWF, sliding window strategies are concentrated, where we mathematically show how states are estimated at the next window $t+1$.

In the following analysis, suppose that $x_m$ and its measurements are to be removed from window in both SWO and SWF. And $x_b$ denotes all direct neighboring parameters of $x_m$, also known as Markov blanket of $x_m$ (see Fig. 2).

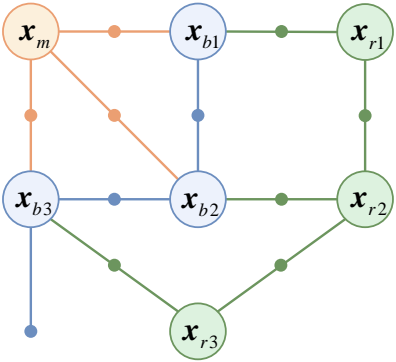

**Fig. 2.** Illustration of the Markov blanket of $x_m$ which will be removed from window. Edges denote observations between nodes and nodes denote the system states. $x_m$ (orange) is directly connected to $x_b = \begin{bmatrix} x_{b1}^T & x_{b2}^T & x_{b3}^T \end{bmatrix}^T$ by observations. Thus, $x_b$ (blue) is the Markov blanket of $x_m$. Other states are denoted by $x_r$ (green).

### 4.1 Sliding Window Operation in SWO

SWO utilizes all measurements to estimate states in the current window, but it only converts measurements of $x_m$ into the posteriori (see Fig. 3). Thus, SWO only considers measurements connected to $x_m$, denoted by $Z_{t,m}$, when window slides, which can be expressed by

$$\underbrace{\left(P_{t|t-1}^{-1} + \sum_{i=t}^{t+s_{t,m}} h_i^T \Lambda_i h_i\right)}_{N_{t,m}} \delta \hat{X}_{t,m} = \underbrace{P_{t|t-1}^{-1} \delta X_{t|t-1} + \sum_{i=t}^{t+s_{t,m}} h_i^T \Lambda_i l_i}_{b_{t,m}}. \tag{40}$$

The superscript of iteration number $i$ is omitted for simplicity. Note that measurements in $Z_{t,m}$ connect $x_m$ and $x_b$, but are irrelevant to $x_r$, so that $N_{t,m}$ has the following form:

$$N_{t,m} = \begin{bmatrix} N_{mm} & N_{mb} \\ N_{mb}^T & N_{bb} \end{bmatrix}. \tag{41}$$

In order to remove $x_m$ from window without information loss, marginalization is carried out by Schur complement over the Markov blanket [53]:

$$\begin{bmatrix} N_{mm} & N_{mb} \\ N_{mb}^T & N_{bb} \end{bmatrix} \begin{bmatrix} \delta x_{t,m} \\ \delta x_{t,b} \end{bmatrix} = \begin{bmatrix} b_m \\ b_b \end{bmatrix}$$
$$\Rightarrow \underbrace{\left(N_{bb} - N_{mb}^T N_{mm}^{-1} N_{mb}\right)}_{N_{t,b}^{-1}} \delta x_{t,b} = \underbrace{b_b - N_{mb}^T N_{mm}^{-1} b_m}_{b_{t,b}}. \tag{42}$$

Considering

$$\delta \hat{x}_{t,b} = N_{t,b}^{-1} b_{t,b}, \tag{43}$$

variable $x_{t,b}$ is constrained by $\delta \hat{x}_{t,b}$ and $N_{t,b}^{-1}$ in the window $t+1$.

Additionally, remind that $N_{t,b}^{-1}$ is dependent on linearization points of $x_m$ and $x_b$ due to nonlinearity, where it is evaluated at the estimates $\hat{X}_t$. Furthermore, $N_{t,b}^{-1}$ consists of measurements of $x_m$, which have been removed. Thus, $N_{t,b}^{-1}$ is unable to be updated with the newest estimates at window $t+1$, i.e. linearization points are fixed. However, $x_b$ is still estimated in the next window, which leads to linearization points of the measurement Jacobian matrix and residual vector changed. Thereafter, linearization points of the same variables are different compared the priori with the Jacobi and residuals, resulting in wrong observability of the system [54]. Thus, linearization points should remain the fixed instead of the updated, known as first estimate Jacobian (FEJ). Obviously, FEJ may lead to large linearization error, which can be represented by the difference between current estimates and fixed linearization points of the variable $x_b$:

$$\Delta x_{t+1,b} = \hat{x}_{t+1,b} - \hat{x}_{t,b}. \tag{44}$$

Fortunately, the first order of this linearization error can be compensated by regarding $b_{t,b}$ as a function of $x_b$ [38], which has the following form

$$\Delta b_{t,b} = N_{t,b} \Delta x_{t+1,b}. \tag{45}$$

It shows that $\Delta x_{t+1,b}$ is also constrained by $N_{t,b}^{-1}$. Then, combining (43), $\delta \hat{x}_{t,b} + \Delta x_{t+1,b}$ and $N_{t,b}^{-1}$ can be regarded as a virtual measurement (46) with $w_{t,p} \sim \mathcal{N}\left(0, N_{t,b}^{-1}\right)$.

$$\delta z_{t,p} = \delta \hat{x}_{t,b} + \Delta x_{t+1,b} + w_{t,p}. \quad (46)$$

Note that only $x_b$, the Markov blanket of $x_m$, is constrained in the window $k+1$ and other unconstrained variables are theoretically equivalent to being constrained by infinite covariance. Therefore, these constraints can be expressed by

$$P_{t+1|t} = \begin{bmatrix} N_{t,b}^{-1} & & \\ & \infty I & \\ & & \infty I \end{bmatrix}, \delta X_{t+1|t} = \begin{bmatrix} \delta \hat{x}_{t,b} + \Delta x_{t+1,b} \\ 0 \\ 0 \end{bmatrix}. \quad (47)$$

Where, variables in the next window are defined by

$$X_{t+1} = \begin{bmatrix} x_b^T & x_r^T & x_n^T \end{bmatrix}^T, \quad (48)$$

and $x_n$ denotes newly coming variables after window slides.

By utilizing measurements in $Z_{t+1}$ and considering the virtual measurements, variables in the window $k+1$ can be optimized by (49) expressed with IFR.

$$\left( P_{t+1|t}^{-1} + \sum_{i=t+1}^{t+s_{t+1}} h_i^T \Lambda_i h_i \right) \delta \hat{X}_{t+1} = P_{t|t}^{-1} \delta X_{t+1|t} + \sum_{i=t+1}^{t+s_{t+1}} h_i^T \Lambda_i l_i. \quad (49)$$

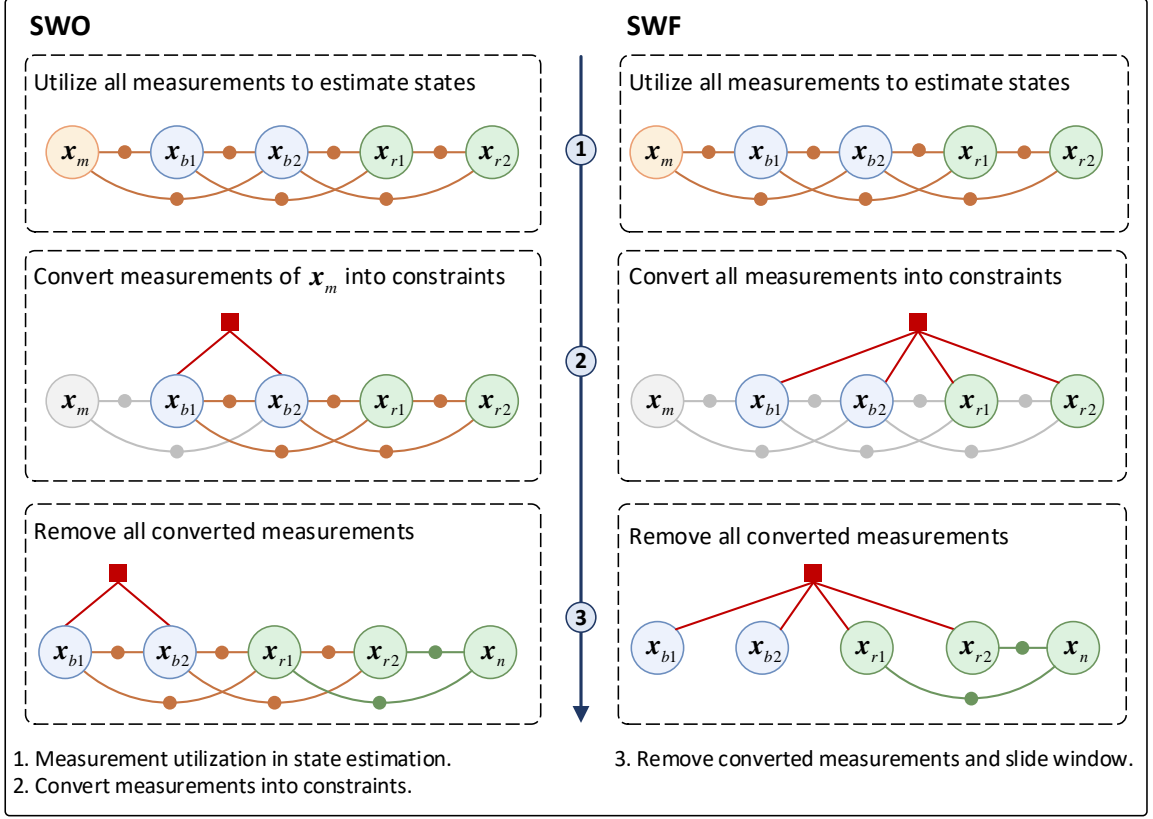

**Fig. 3.** Illustration of sliding window strategy in SWO and SWF. All measurements in the window are utilized (brown) to optimize states in both SWO and SWF. Differently, SWO only converts measurements of $x_m$ into constraints and then removes them (grey), which constrains $x_b$ (red). And the rest measurements can be utilized in the next window. SWF converts all utilized measurements into the priori, which constrains $x_b$ and $x_r$ in the next window. And thus, all these converted measurements are removed. With FEJ strategy taken into consideration, linearization points of $x_r$ are early fixed in SWF compared with SWO. $x_n$ denotes newly coming states after window slides.

## 4.2 Sliding window operation in SWF

In SWF, all measurements in the window are used to update states and then are converted into the posteriori (50), which is different from SWO, as shown in Fig. 3.

$$\delta \hat{X}_t = \delta X_{t|t-1} + K_t \left( L_t - H_t \delta X_{t|t-1} \right),$$
$$P_t = \left( I - K_t H_t \right) P_{t|t-1}. \quad (50)$$

Considering $P_t = N_t^{-1}$, $P_t$ has the following form by mathematically inverting $N_t$ (see Appendix):

$$P_t = N_t^{-1} = \begin{bmatrix} N_{mm} & N_{mb'} \\ N_{mb'}^T & N_{b'b'} \end{bmatrix}^{-1}$$
$$= \begin{bmatrix} \left( N_{mm} - N_{mb'} N_{b'b'}^{-1} N_{mb'}^T \right)^{-1} & C \\ C^T & \left( N_{b'b'} - N_{mb'}^T N_{mm}^{-1} N_{mb'} \right)^{-1} \end{bmatrix}, \quad (51)$$

where,

$$N_{b'b'} = \begin{bmatrix} N_{bb} & N_{br} \\ N_{br}^T & N_{rr} \end{bmatrix}, N_{mb'} = \begin{bmatrix} N_{mb} & N_{mr} \end{bmatrix}. \quad (52)$$

When window slides, $x_m$ with measurements in $Z_t$ is removed from window, and covariance of $x_m$ is directly deleted. Note that estimated error state $\delta \hat{X}_t$ should be compensated back to the propagated value via

$$\hat{X}_t = X_{t|t-1} - \delta\hat{X}_t. \tag{53}$$

Thus, current estimates $\hat{X}_t$ are regarded as zero error and then $\delta\hat{X}_t$ is set to zero vector after compensation, which is the closed-loop system. Thereafter, $x_b$ and $x_r$ are constrained by

$$N_{t,b'}^{-1} = \left(N_{b'b'} - N_{mb'}^T N_{mm}^{-1} N_{mb'}\right)^{-1}, \delta\hat{x}_{t,b'} = \begin{bmatrix} \delta\hat{x}_b \\ \delta\hat{x}_r \end{bmatrix} = \mathbf{0}. \tag{54}$$

Similarly, linearization points of $x_b$ and $x_r$ should remain the fixed instead of the updated according to FEJ strategy. However, the leading increasing linearization error is not compensated in SWF. Thus, the priori is denoted by (55) with $w'_{t,p} \sim \mathcal{N}\left(\mathbf{0}, N_{t,b'}^{-1}\right)$

$$\delta z'_{t,p} = \delta\hat{x}_{t,b'} + w'_{t,p}. \tag{55}$$

Then, constraints in the window $k+1$ can be expressed by

$$P'_{t+1|t} = \begin{bmatrix} \left(N_{b'b'} - N_{mb'}^T N_{mm}^{-1} N_{mb'}\right)^{-1} & \\ & \infty I \end{bmatrix}, \delta X'_{t|t} = \begin{bmatrix} \delta\hat{x}_{t,b'} \\ \mathbf{0} \end{bmatrix}. \tag{56}$$

Consequently, incorporating virtual measurements (55), variables in the window $k+1$ can be estimated by

$$\delta\hat{X}_{t+1} = K'_{t+1} L'_{t+1},$$
$$K'_{t+1} = P'_{t+1|t} H'^T_{t+1} \left(\Lambda^{-1}_{t+1} + H'^T_{t+1} P_{t|t-1} H'^T_{t+1}\right)^{-1} \tag{57}$$

Where, $H'_{t+1}$ and $L'_{t+1}$ are corresponding Jacobian matrix as well as residual vector using measurements in $Z'_{t+1}$ of SWF, where the two measurement sets of SWF and SWO are different from each other and will be discussed in the following.

## 4.3 Discrepancy Analysis Between SWO and SWF

In order to analyze discrepancies between SWO and SWF, expression of state estimation using SWF in window $t+1$ (57) is rewritten as

$$\left(P'_{t+1|t} + \sum_{i=t+1}^{t+s'_{t+1}} h'^T_i \Lambda_i h'_i\right) \delta\hat{X}'_{t+1} = \sum_{i=t+1}^{t+s'_{t+1}} h'^T_i \Lambda_i l'_i. \tag{58}$$

Then, comparing (49) with (58), it is obviously shown that variables estimated by SWO and SWF in the window $t+1$ exist discrepancies. Since they are equivalent at timestamp $t$, the reasons of these discrepancies can be attributed to several aspects caused by sliding window strategy.

1)  **Different Prior constraints.** From (42), SWO only converts measurements of $x_m$ using marginalization, which constrains $x_b$ only. Moreover, it considers full prior constraints, including both $P^{-1}_{t+1|t}$ and $\delta X_{t+1|t}$ in the window $t+1$. On the contrary, SWF constrains $x_b$ and $x_r$ in the window $t+1$ (54), since it converts all measurements in $Z_t$ into the priori. Besides, note that $\delta\hat{x}_{t,b'}$ is set to zero vector (54), and the priori satisfies

$$N_{t,b'} \delta\hat{x}_{t,b'} = b_{t,b'}. \tag{59}$$

Thus, $b_{t,b'}$ is a zero vector as well. However, $b_{t,b}$ is obtained by marginalization in SWO (43), which is set to a nonzero vector and constrains variables in the window $t+1$. As a result, zero vector $\delta\hat{x}_{t,b'}$ is equivalent to ignoring part of constraints compared with SWO.

2)  **Different Measurement Utilization.** All measurements converted to the constraints should be removed in both SWO or SWF. Otherwise, measurements will be repeatedly used. Theoretically, prior constraints $P^{-1}_{t+1|t}$ in SWO can be rewritten as:

$$P^{-1}_{t+1|t} \Leftrightarrow P^{-1}_{t|t-1} + \sum_{i=t}^{t+s_t} h^T_i \Lambda_i h_i. \tag{60}$$

Then the information matrix in the window $t+1$ can be expressed by

$$N_{t+1} \Leftrightarrow P^{-1}_{t|t-1} + \sum_{i=t}^{t+s_t} h^T_i \Lambda_i h_i + \sum_{i=t+1}^{t+s_{t+1}} h^T_i \Lambda_i h_i. \tag{61}$$

Suppose $Z_t \cap Z_{t+1} \neq \varnothing$, some measurements will be repeatedly contributed to the information matrix, leading to the wrong posteriori. Then, this further results in wrong constraints and suboptimal solution in the subsequent estimation, especially fusing measurements from multiple sources.

Consequently, measurements converted into the constraints are no longer utilized in window $t+1$. SWO only converts measurements in $Z_{t,m}$ to the priori, leading to $Z_t \cap Z_{t+1} = Z_{t,r}$ without repeatedly use. Conversely, SWF converts and removes all measurements in $Z_t$, and thus $Z'_{t+1}$ in SWF is different from $Z_{t+1}$ in SWO, which satisfies $Z_t \cap Z'_{t+1} = \varnothing$. In other words, measurements in $Z_{t,r}$ are re-linearized compared with SWF, which is the reason that SWO outperforms SWF.

3)  **Different FEJ Compensation.** Linearization points of constrained states should be fixed according to FEJ strategy, where the linearization error may be increasingly larger and the first order of this linearization error can be compensated in SWO (47). However, it is ignored in SWF.

## 4.4 Applications of SWO and SWF in Visual SLAM: Two Representative Frameworks

Strategies in two representative and state-of-the-art schemes are further discussed based on visual inertial odometry: VINS-mono and MSCKF, which utilizes SWO and SWF respectively. Since the two schemes apply different model of inertial measurement unit (IMU), visual measurements processing is focused in state estimation.

1)  *VINS-mono:* VINS-mono utilizes landmarks, which have been tracked by at least two frames, to estimate states in the window. After estimation, the oldest frame is to be removed. Next, the Markov blanket of the oldest frame is built, which contains landmarks observed by it, i.e. $x_{l,1}$ and $x_{l,2}$ in this case. Marginalization is then carried out according to (42), resulting in constraints of the Markov blanket in the next window. Since only measurements of the removed frame are converted to the posteriori, others remain available in the next window due to marginalization.

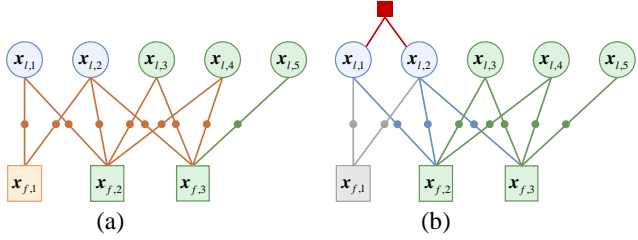

**Fig. 4.** An example illustrates measurement utilization and sliding window strategy in VINS-mono. All tracked landmarks are jointly estimated with frames by all measurements (a). Only measurements of the oldest frame are removed, and thus the Markov blanket of $x_{f,1}$ which consists of $x_{l,1}$ and $x_{l,2}$ is constrained (b).

*2) MSCKF:* Landmarks will be used if one of the following conditions are satisfied in MSCKF, where strategies utilized in it are simplified for discussion, as shown in Fig. 5.

i.Landmarks are no longer observed by latest frame;

ii.Landmarks are observed by at least three frames.

In this case, $x_{l,1}$ and $x_{l,2}$ are utilized, which satisfies the two conditions respectively. With measurements connected to the two landmarks utilized to estimate states, these measurements are all converted into constraints. Then, these measurements as well as the oldest frame, $x_{f,1}$, are removed from window. Since the two landmarks are no longer observed, they are also removed. Thus, $x_{f,2}$ and $x_{f,3}$ are constrained in the next estimation. And measurements are utilized only once.

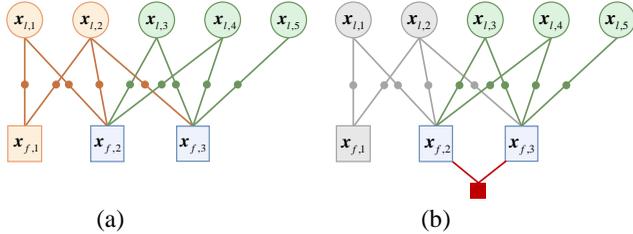

**Fig. 5.** An example illustrates measurement utilization and sliding window strategy in MSCKF. Only satisfied landmarks (brown) are used to update states (a). And all these used measurements are removed and converted to the priori after estimation, which constrains $x_{f,2}$ and $x_{f,3}$ (b).

Comparing the two applications of SWO with SWF in visual odometry, discrepancies between SWO and SWF will show up even from the first window, since different measurement utilization strategies are applied in SWO and SWF, and which further leads to different constraints.

*Proposition 3:* Theoretically, discrepancies are caused by sliding strategy instead of state estimation algorithms, where the estimates are equal at the beginning but SWO outperforms SWF in the subsequent estimation. When window slides, SWO utilizes marginalization over the Markov blanket (42) converting part of measurements, which enables re-linearization. SWF converts all used measurements into the priori and ignores part of constraints (54). Due to different measurements converted to the priori, measurement utilization is further different from each other. Besides, linearization error in SWO caused by FEJ can be compensated (45), while it is increasingly larger in SWF (55).

## 5 Strategy Modification On Sliding window

### FILTERING

In this section, strategies of existing SWF are adjusted in order to achieve the same performance in accuracy with SWO. Measurement set $Z_t$ is firstly divided into two parts, where they contain measurements of the removing states and the others respectively. A 2-step update strategy, ensuring the same measurement utilization and constraints, is utilized which makes SWF equivalent to SWO in accuracy.

#### 5.1 Observation Preparation

Comparing (42) with (54), it is shown that marginalization using Schur complement in SWO shares the same mathematical form with directly deleting covariance in SWF. In other words, the prior constraints will be equivalent once the same measurements are converted. Thereafter, the measurement utilization in the next window is equal as well.

So, given the estimating states $X_t = \begin{bmatrix} x_t & \cdots & x_{t+N} \end{bmatrix}$ with measurements contained in set $Z_t$, the to-be-removed states $x_m$ should be firstly selected. Then, divide $Z_t$ into $Z_{t,m}$ and $Z_{t,r}$, which denotes set of measurements observed by $x_m$ and the rest respectively, where the two sets satisfy

$$Z_{t,m} \cup Z_{t,r} = Z_t, \ Z_{t,m} \cap Z_{t,r} = \varnothing. \tag{62}$$

#### 5.2 2-step Update

Since the final estimates of sequential update equal to the solution of batch estimation according to proposition 1, measurements in $Z_{t,m}$ and $Z_{t,r}$ are updated sequentially in order to ensure the equivalent estimates. Then, only measurements in $Z_{t,m}$ are converted into constraints in the next window, ensuring the same priori. Additionally, with the IFR expression in (49), it is demonstrated that the ignored constraints $\delta X_{t|t-1}$ and FEJ compensation $\Delta X_t$ can also be taken into consideration in SWF equivalently. Thus, after attaining the prior constraints $P_{t|t-1}$ and $\hat{X}_{t|t-1}^{(0)}$, the 2-step update strategy in $i$-th iteration is adjusted as follows (see Fig. 6):

**Step 1:** Update states and covariance using measurements in $Z_{t,m}$ with full information incorporated and FEJ compensation. Evaluate Jacobian matrix at $\hat{X}_{t|t-1}^{(i)}$, which is the estimates of previous iteration. Note that the estimates, $\delta \hat{X}_{t,m}^{(i)}$, should not be compensated to states. Otherwise, the consistency of Jacobi will be broken, and wrong observability will be introduced due to different linearization points applied in the same states.

**Step 2:** Sequentially update states by measurements belonging to $Z_r$ regarding the estimates of step 1 as a virtual measurement. Similarly, evaluate Jacobian matrix at $\hat{X}_{t|t-1}^{(i)}$ which ensures the Jacobian consistency. Note that $\delta \hat{X}_{t,m}^{(i)}$ and $P_{t,m}^{(i)}$, should be used as virtual measurements, in order to utilize measurement information in $Z_{t,m}$, as the stated sequential update in proposition 2. The covariance matrix in this step is not updated to ensure the same measurements converted to posteriori.

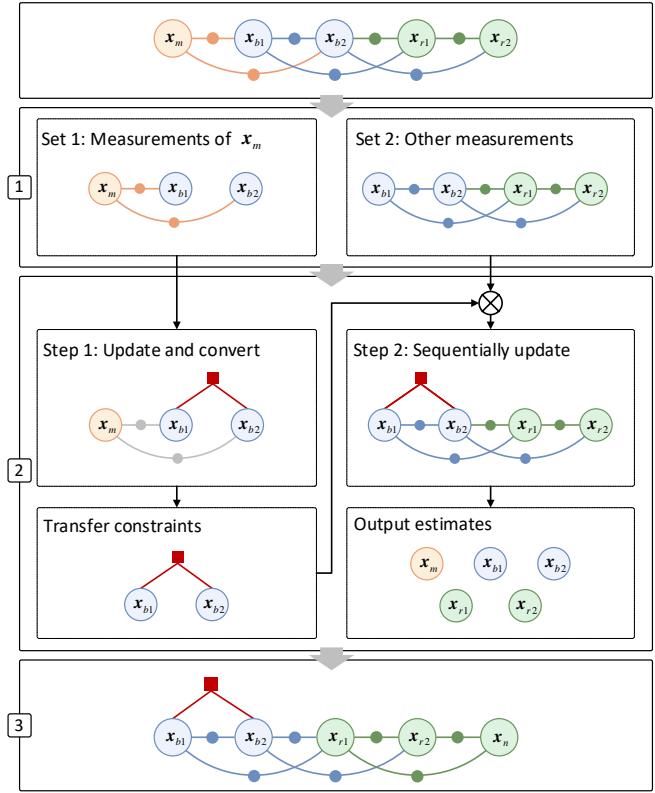

**Fig. 6.** Overview of the strategy adjusted SWF. Suppose variables ( $\boldsymbol{x}_m$ , $\boldsymbol{x}_b$ , $\boldsymbol{x}_r$ ) in the current window are to be estimated. Three procedures are executed in the strategy adjusted SWF. 1) Divide measurements into two sets, where the first set contains measurements of $\boldsymbol{x}_m$ and the other contains the rest ones. 2) Adjust SWF by a 2-step update strategy. Step 1 utilizes measurements in set 1 to update and convert them into the constraints. Step 2 sequentially updates states with constraints from step 1 incorporated, then output the estimates. 3) Slide the window and discard measurements of $\boldsymbol{x}_m$ . Thus, the 2-step update strategy enables the equivalent prior constraints, measurement utilization, and FEJ compensation compared with SWO.

Step 1 converts the same measurements compared with SWO, ensuring the equivalent constraints in the next window. On the other hand, step 2 fully utilizes all available measurements without converting them. Then, other measurements can be still used (re-linearization) in the subsequent window, which ensures the same measurement utilization and thus the same estimates as SWO in the subsequent. The whole procedure of the 2-step update can be referred to algorithm 2.

---

**Algorithm 2** Strategy adjusted sliding window filtering

set $\delta\boldsymbol{X}_{t|t-1}^{(0)} \leftarrow \delta\boldsymbol{X}_{t|t-1}, i \leftarrow 1, MaxIter, \boldsymbol{P}_{t|t-1}$

divide $\boldsymbol{Z}_k$ into $\boldsymbol{Z}_{k,m}, \boldsymbol{Z}_{k,r}$ , $s.t. \boldsymbol{Z}_{t,m} \bigcup \boldsymbol{Z}_{t,r} = \boldsymbol{Z}_t, \boldsymbol{Z}_{t,m} \bigcap \boldsymbol{Z}_{t,r} = \varnothing$

**while** $i < MaxIter$

    // update and convert measurements in $\boldsymbol{Z}_{t,m}$

    $\delta\hat{\boldsymbol{X}}_{t,m}^{(i)} \leftarrow \delta\boldsymbol{X}_{t|t-1}^{(i-1)} + \Delta\boldsymbol{X}_k$

    $+\boldsymbol{P}_{t|t-1}^{-1}\boldsymbol{H}_{t,m}^{(i)\,T}\boldsymbol{\Lambda}_{t,m}\left(\boldsymbol{L}_{t,m}^{(i)} - \boldsymbol{H}_{t,m}^{(i)}\left(\delta\boldsymbol{X}_{t|t-1}^{(i-1)} + \Delta\boldsymbol{X}_t\right)\right)$

---

$\boldsymbol{P}_{t,m}^{(i)} \leftarrow \left(\boldsymbol{I} - \boldsymbol{K}_{t,m}^{(i)}\boldsymbol{H}_{t,m}^{(i)}\right)\boldsymbol{P}_{t|t-1}$

// sequentially update measurements in $\boldsymbol{Z}_{t,r}$

$\delta\hat{\boldsymbol{X}}_{t,r}^{(i)} \leftarrow \delta\hat{\boldsymbol{X}}_{t,m}^{(i)} + \boldsymbol{P}_{t,m}^{(i)}\boldsymbol{H}_{t,r}^{(i)\,T}\boldsymbol{\Lambda}_{t,r}\left(\boldsymbol{L}_{t,r}^{(i)\,T} - \boldsymbol{H}_{t,r}^{(i)}\delta\hat{\boldsymbol{X}}_{t,m}^{(i)}\right)$

$\hat{\boldsymbol{X}}_{t|t-1}^{(i)} \leftarrow \hat{\boldsymbol{X}}_{t|t-1}^{(i-1)} - \delta\hat{\boldsymbol{X}}_{t,r}^{(i)}$

**if** converge

    // transfer constraints converted by $\boldsymbol{Z}_{t,m}$

    $\boldsymbol{P}_t \leftarrow \boldsymbol{P}_{t,m}^{(i)}, \delta\hat{\boldsymbol{X}}_{t,m}^{(i)}$

    // return estimates updated by all measurements

    $\hat{\boldsymbol{X}}_t \leftarrow \hat{\boldsymbol{X}}_{t|t-1}^{(i)}$

    **return** $\hat{\boldsymbol{X}}_t, \boldsymbol{P}_t, \delta\hat{\boldsymbol{X}}_{t,m}^{(i)}$

**end if**

$i \leftarrow i + 1$

**end while**

**output estimates** $\hat{\boldsymbol{X}}_t$

**slide window**

**transfer constraints** $\boldsymbol{P}_{t+1|t} \leftarrow \left[\!\left[\boldsymbol{P}_t\right]\!\right]_{\boldsymbol{x}_m}, \delta\boldsymbol{X}_{t+1|t} \leftarrow \left[\!\left[\delta\boldsymbol{X}_{t,m}^{(i)}\right]\!\right]_{\boldsymbol{x}_m}$

---

Where, operator $\llbracket \boldsymbol{v} \rrbracket_{\boldsymbol{x}}$ denotes delete elements related to $\boldsymbol{x}$ in matrix or vector $\boldsymbol{v}$ . Thus, the strategy adjusted SWF has the same prior constraints, measurement utilization, and FEJ compensation.

*Proposition 4:* By adjusting existing strategy to the 2-step update strategy, which ensures the same measurements utilized and converted to the constraints, strategy adjusted SWF mathematically equals to SWO. It again demonstrates that the discrepancies between SWF and SWO are caused by strategies instead of state estimation algorithms.

# 6 EXPERIMENTAL VALIDATION BASED ON VISUAL SLAM

In this section, a series of experiments are presented based on visual SLAM to validate the theoretical analysis, in particular, proposition 2-4. Specifically, Monte-Carlo experiments are conducted first to validate the propositions. And then, the strategy adjusted SWF (SWF-SA) is examined by ablation experiments based on KITTI dataset. Besides, assume that there are no external dynamic or other constraints information in all these experiments, where $\boldsymbol{\Phi}$ is set to the identity. Since SLAM is a highly nonlinear system, iteration strategy has to be employed, where the threshold in algorithm 1 and algorithm 2 is also set equally. The weight of each visual measurement is equally set to 2 pixels. Other strategies, such as outlier rejection, are implemented equivalently in the two estimators as well. In our implementations, IFR in optimization-based methods, while KFR is utilized in filtering-based counterparts.

The root mean square error (RMSE) and the discrepancy are applied to evaluate experimental results. Discrepancy is defined as the difference between estimates of the filtering-based and optimization-based approaches, then the Euclidean norm is calculated.

## 6.1 Monte-Carlo experiments

A series of Monte-Carlo experiments are conducted, includ-ing 1) solve camera poses as well as landmarks by batch estimation and sequential update, 2) apply real-time operation using SWO and SWF, and 3) adjust existing strategy in SWF then compare with SWO. Each experiment processes the same data of 50 Monte Carlo trials, where the camera moves along a circular path, as shown in Fig. 7. Initial values of frames and landmarks for linearization are provided for simplicity, rather than triangulate feature points and determine camera poses using structure-from-motion, which will not affect the equivalence between estimators.

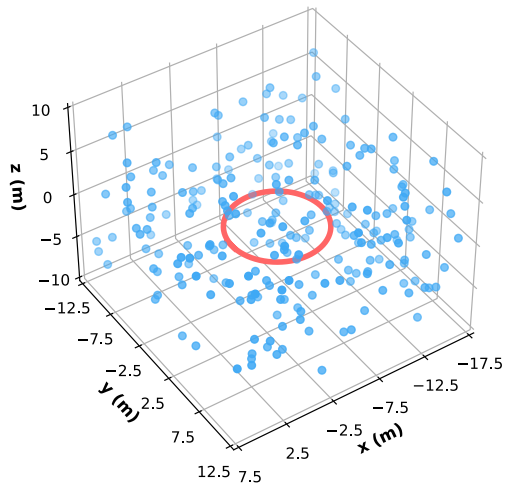

**Fig. 7.** A stereo camera is simulated, moving along the circular trajectory (red) with landmarks (blue) randomly generated in the space.

### 1) *Equivalence between batch estimation and sequential update.*

The system states of each trial include all camera poses as well as all landmarks observed by at least 2 frames. Then, all tracked feature points are utilized as measurements in both approaches. Discrepancies between the two methods at arbitrary timestamps can be obtained, where Fig. 8 shows averaged discrepancies between batch estimation and the sequential update at each timestamp, demonstrating the equivalence between the two estimators if measurements, observation model, stochastic model, and linearization points are equivalently utilized. This slight difference may be introduced by computational error.

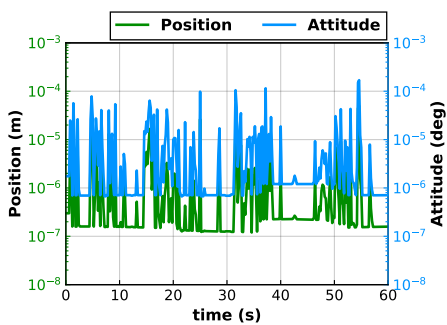

**Fig. 8.** Equivalence between batch estimation and sequential update based on 50 Monte Carlo trials. The maximum discrepancies of position and attitude are no more than $10^{-4}\,m$, $10^{-3}$ deg respectively, which demonstrates the equivalence between batch estimation and sequential update.

### 2) *Discrepancies between SWO and existing SWF.*

In our implementation of SWO and SWF, the length of window is set to 20 frames as well as landmarks observed by them, and all tracked feature points are used to update states, ensuring the same measurements utilization at the beginning. Thus, estimating results of the first frame will be equivalent (see Fig. 9). Differently, SWO adopts marginalization to convert measurements of the first frame to priori. On the contrary, SWF converts all utilized measurements into constraints and then discards them. So, this equivalence corrupts from the second window due to different measurements converted, which further leads to different measurement utilization. And thus existing SWF is inequal to SWO.

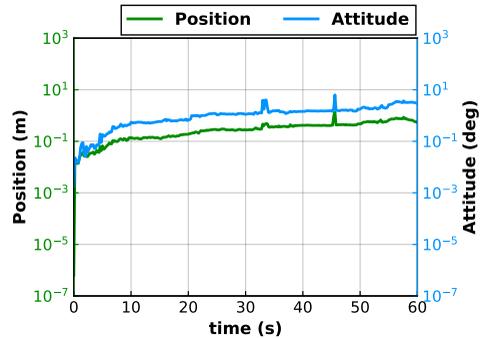

**Fig. 9.** Averaged discrepancies between SWO and existing SWF, where SWO and SWF are equal ($10^{-7}\,m$ and $10^{-7}$ deg respectively) at beginning due to the same measurement utilization and prior knowledge. This equivalence corrupts due to existing strategies applied in SWF, which intuitively shows the inequality between SWO and SWF.

### 3) *Equivalence between SWO and SWF-SA.*

Strategies in SWF are adjusted, by a 2-step update strategy stated in proposition 4, which ensures the same measurement utilization and prior constraints with full constraints and FEJ compensation incorporated. Thus, discrepancies between the optimization and filtering are disappeared, remaining computational error, demonstrating the equivalence between SWO and SWF-SA, as shown in Fig. 10.

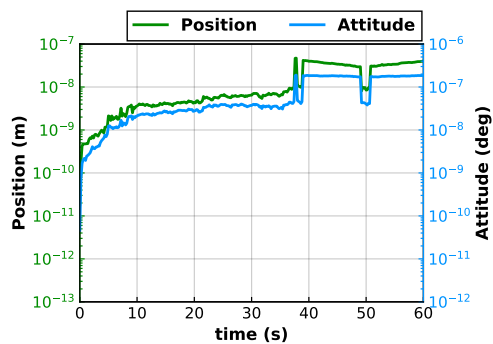

**Fig. 10.** Averaged discrepancies between SWO and SWF-SA based on 50 Monte Carlo trials. The maximum discrepancies of position and attitude are no more than $10^{-6}\,m$, $10^{-7}$ deg respectively, showing the equivalence between SWO and SWF-SA.

Finally, Fig. 11 shows RMSE of each trial, which demonstrates that SWO indeed significantly performs better than existing SWF due to strategies, specifically re-linearization. But with a 2-step modification, the SWF-SA achieves the same estimating accuracy with SWO. Besides, batch estimation fully utilizes measurement information, which shows the best estimating accuracy in most trials compared with sliding window estimation and is consistent with the preceding analysis.

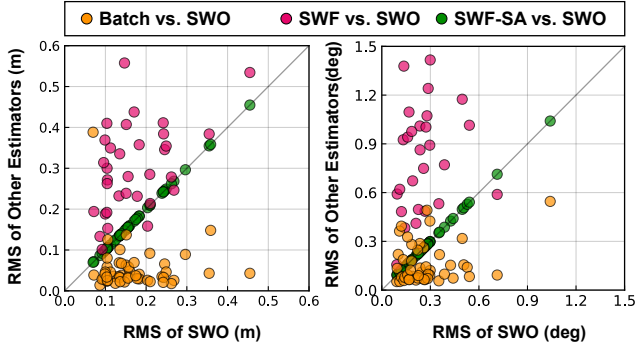

**Fig. 11.** Comparison of Monte-Carlo experimental results. Each circle represents RMSE of once trial and the color denotes the comparing group.

## 6.2 Ablation studies based on KITTI

In this section, effects of incomplete constraints and FEJ compensation in SWF are analyzed through ablation experiments based on KITTI dataset [52]. In the ablation experiments, two factors are analyzed, ignored constraints $\delta\hat{x}_{t,b}$ and FEJ compensation. Since SWO as well as SWF-SA achieves the best estimates in real-time operation from the simulation experiments, SWO is conducted as the control group and SWF-SA is firstly compared with it. Then, based on SWF-SA, the first factor is analyzed by ignoring $\delta\hat{x}_{t,b}$, while strategy of FEJ compensation remains equivalent to SWF-SA, namely SWF-FEJ. Finally, linearization error caused by FEJ is not compensated but with full priori incorporated, analyzing the second factor, namely SWF-Full. Note that other strategies keep equivalent to SWF-SA, including measurement utilization, outlier removal, et al. And ORBSLAM 2 [55] is utilized as our frontend to provide initial values of camera frames and landmarks.

Comparing with SWO, TABLE I gives the statistics of discrepancies of the ablation experiments. The maximum reflects the extreme discrepancies but it may be affected by wrong estimates. The average can reflect the general discrepancy over each sequence. First, SWF-SA is strictly equivalent to SWO, where both the maximum and average are no more than $10^{-6}$ on both translation and attitude, which is negligible during state estimation and is consistent with our previous analysis. Apparently, both $\delta\hat{x}_{t,b}$ and FEJ compensation significantly affect equivalence between SWO and SWF. Specifically, SWF-FEJ shows better equivalence compared with SWF-Full in general, which indicates that compensating linearization error caused by FEJ is necessary in a high nonlinear system. Meanwhile, incomplete constraints may lead to large extreme value on attitude, which is indispensable during state estimation as well.

It is important to note that, in implementation of MSCKF 2.0 [56], correct observable structure of the estimator using FEJ is more important than linearization error introduced by FEJ, since they utilize inertial measurement unit to provide reliable initial value. In our experiments, linearization points are obtained from a pure visual scheme, where the linearization error may be larger compared with the visual inertial system. Thus, for a high nonlinear system without precise initial value, the effect of linearization error may be greater than incomplete prior constraints.

TABLE I
STATISTICS OF DISCREPANCY COMPARED WITH SWO ON ABLATION EXPERIMENTAL RESULTS

| Sequence | Translation ( $\log_{10} m$ ) | | | | | | Attitude ( $\log_{10}$ deg ) | | | | | |
| | SWF-SA | | SWF-FEJ | | SWF-Full | | SWF-SA | | SWF-FEJ | | SWF-Full | |
| | max | mean | max | mean | max | mean | max | mean | max | mean | max | mean |
| 00 | -7.98 | -8.62 | 1.13 | 0.03 | **1.21** | **0.15** | -6.68 | -8.51 | 1.40 | 0.32 | **1.77** | **0.59** |
| 01 | -8.38 | -9.14 | **1.59** | 0.72 | 1.52 | **0.82** | -8.28 | -9.73 | **1.06** | -0.14 | 0.79 | **0.02** |
| 02 | -7.72 | -8.60 | **0.78** | **0.21** | 0.77 | 0.21 | -7.43 | -8.52 | **1.47** | 0.34 | 1.17 | **0.45** |
| 03 | -8.50 | -9.08 | 1.37 | **0.27** | **1.37** | 0.16 | -8.47 | -9.24 | 1.42 | 0.01 | **1.42** | **0.15** |
| 04 | -8.87 | -9.17 | 0.48 | -0.01 | **0.74** | **0.21** | -8.85 | -9.33 | **0.77** | **0.43** | 0.54 | 0.11 |
| 05 | -8.12 | -8.58 | 0.67 | **0.30** | **1.10** | 0.21 | -7.73 | -8.54 | 1.14 | **0.56** | **1.38** | 0.46 |
| 06 | -8.32 | -9.07 | 0.85 | 0.23 | **1.50** | **0.26** | -7.80 | -8.83 | **1.32** | **0.40** | 1.10 | 0.36 |
| 07 | -7.35 | -8.55 | 0.50 | 0.12 | **0.69** | **0.12** | -6.90 | -7.71 | **1.20** | **0.46** | 1.16 | 0.31 |
| 08 | -7.56 | -8.66 | 0.88 | 0.23 | **1.53** | **0.30** | -6.74 | -8.31 | **1.44** | 0.51 | 1.37 | **0.58** |
| 09 | -7.88 | -8.84 | 0.60 | **0.14** | **0.91** | 0.13 | -7.54 | -8.72 | 1.20 | 0.36 | **1.39** | **0.37** |
| 10 | -8.18 | -8.69 | 0.96 | 0.11 | **1.00** | **0.16** | -7.85 | -8.33 | 1.19 | 0.57 | **1.35** | **0.59** |

*Note that $\log_{10} 10^x = x$, which shows that statistics in TABLE I reflect the magnitude of discrepancies in ablation experiments.*

Besides, from the simulation analysis between SWO and SWF-SA (Fig. 10), discrepancies between them tend to increase. Nonetheless, KITTI dataset contains large-scale scenarios, where our experimental results intuitively show that by adjusting strategies in existing SWF, the SWF-SA strictly equals to

SWO and discrepancies between them are negligible during estimation (see Fig. 12).

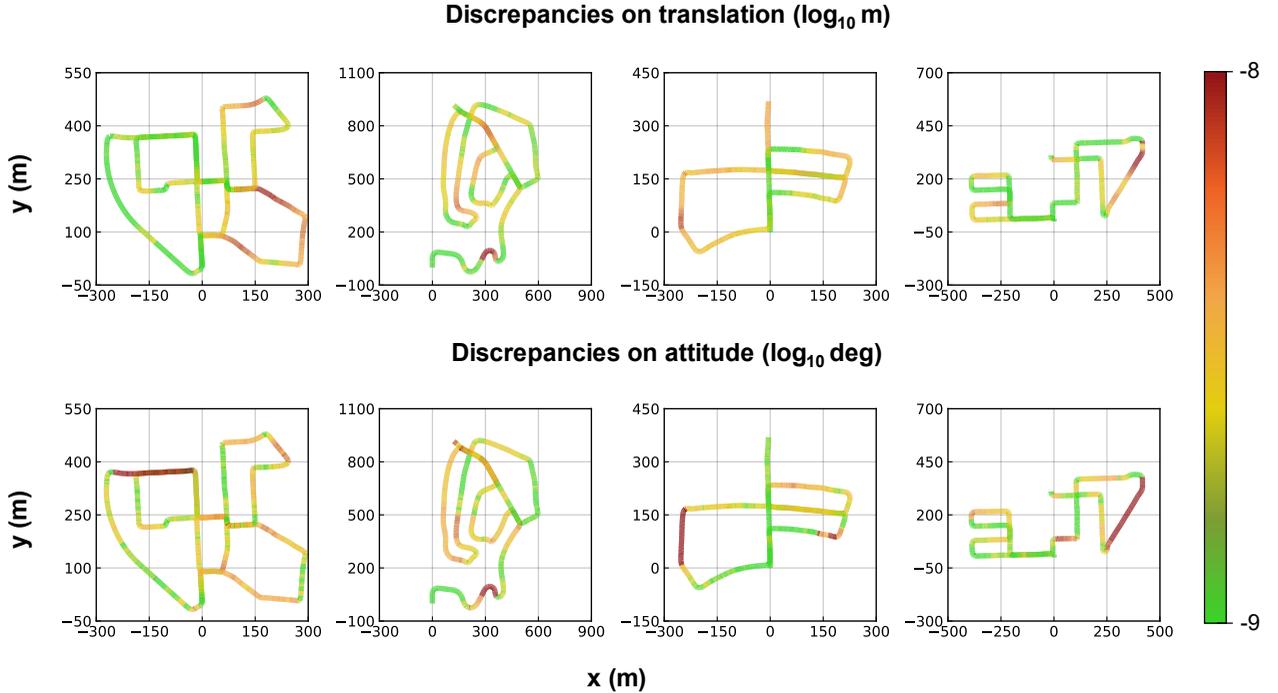

**Fig. 12.** Discrepancies on translations and attitudes between SWF-SA and SWO based on KITTI 00, 02, 05, and 08 sequences. Trajectories are plotted by ground truth, and color represents discrepancies between SWF-SA and SWO.

## 7 CONCLUDING REMARKS

In this paper, discrepancies between optimization-based and filtering-based approaches have been explored by analyzing theories and strategies between batch estimation and sequential update, between SWO and SWF. It demonstrates that the discrepancies are caused by sliding window strategy, where SWF converts different measurements into constraints compared with SWO, leading to different prior constraint and measurement utilization. Additionally, SWF ignores part of constraints and FEJ compensation. Based on previous analysis, a 2-step update strategy is applied in existing SWF, which ensures the same measurements converted, fully incorporates prior constraints and FEJ compensation. The strategy adjusted SWF is evaluated by simulation experiments and ablation experiments based on KITTI dataset, where it is shown that strategy adjusted SWF is strictly equivalent to SWO under the assumption of the same measurements, observation model, stochastic model, and strategies. The discrepancies are caused by different strategies which further leads to different measurement utilization and constraints, instead of state estimation algorithms.

Thus, future research on sensor fusion should thoroughly describe their own algorithms and strategies which essentially affect the estimating results. For example, feature correspondence is main task in computer vision, which should be deeply studied. Then, measurements from the same sensor have different quality, even including outliers, which may lead to fault estimates. For those outliers, they should be carefully detected and removed, while other healthy observations should also be properly weighted by a prior model instead of determining weights through empirical values. Additionally, it is still a challenge to estimate uncertainty of each sensor appropriately for sensor fusion problem, which may lead to wrong estimates as stated before. As for state estimation method, it should be chosen according to dimension of system states as well as number of measurements to ensure efficiency due to different time complexity of the two state estimation methods.

## 8 APPENDIX A

Given a real symmetric positive definite matrix $N$, the inverse of it can be expressed by $N^{-1}$ which satisfies:

$$\underbrace{\begin{bmatrix} N_{mm} & N_{mb} \\ N_{mb}^T & N_{bb} \end{bmatrix}}_{N} \underbrace{\begin{bmatrix} A & B \\ C & D \end{bmatrix}}_{N^{-1}} = \begin{bmatrix} I & \\ & I \end{bmatrix}. \tag{63}$$

Thus, we have:

$$\begin{cases} N_{mm}A + N_{mb}C = I \\ N_{mm}B + N_{mb}D = 0 \\ N_{mb}^T A + N_{bb}C = 0 \\ N_{mb}^T B + N_{bb}D = I \end{cases} \tag{64}$$

Since $N_{mm}$ and $N_{bb}$ are invertible, $N^{-1}$ can be derived by solving (64):

$$\begin{cases} A = \left( N_{mm} - N_{mb}N_{bb}^{-1}N_{mb}^T \right)^{-1} \\ C = B^T = -\left( N_{bb} - N_{mb}^T N_{mm}^{-1}N_{mb} \right)^{-1} N_{mb}^T N_{mm}^{-1}. \\ D = \left( N_{bb} - N_{mb}^T N_{mm}^{-1}N_{mb} \right)^{-1} \end{cases} \tag{65}$$

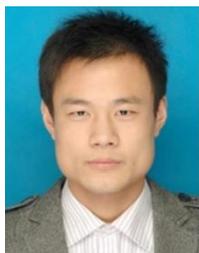

**Feng Zhu** is currently an associate professor at the School of Geodesy and Geomatics, Wuhan University, China. He obtained his B.Sc., Master, and Ph.D. degrees with distinction in Geodesy and Engineering Surveying at the School of Geodesy and Geomatics in Wuhan University in 2012, 2015 and 2019. His current research focuses on multi-sensor integration and its application in position and orientation systems and autonomous systems.

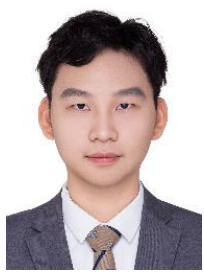

**Zhuo Xu** is currently pursuing his master's degree at the School of Geodesy and Geomatics, Wuhan University, China. He received his B.Sc. degree at the School of Geodesy and Geomatics, Wuhan University in 2022. His research interests include multi-sensor fusion navigation and technology.

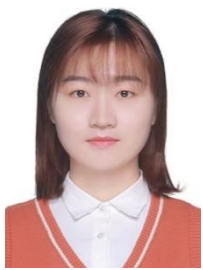

**Xveqing Zhang** is currently pursuing her master's degree at the School of Geodesy and Geomatics, Wuhan University, China. She received her B.Sc. degree at the School of Geodesy and Geomatics, Wuhan University in 2022. Her research interests include multi-sensor fusion navigation technology.

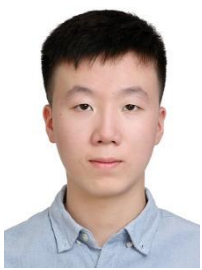

**Yuantai Zhang** is currently pursuing his master's degree at the School of Geodesy and Geomatics, Wuhan University, China. He received his B.Sc. degree from the School of Geodesy and Geomatics, Wuhan University in 2022. His research interests include GNSS, graph optimization and its application in multi-sensor fusion.

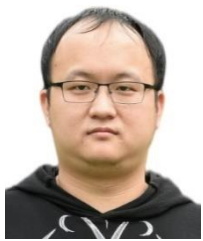

**Weijie Chen** is currently pursuing his Ph.D. degree at the School of Geodesy and Geomatics, Wuhan University, China. He received his B.Sc. and M.Sc. degree at the School of Geodesy and Geomatics, Wuhan University, in 2020 and 2023 respectively. His research interests include multi-sensor fusion navigation and deep learning.

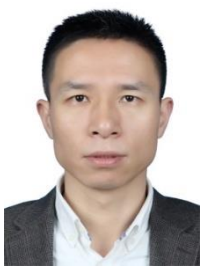

**Xiaohong Zhang** is currently a professor at Wuhan University. He obtained his B.Sc., M.Sc., and Ph.D. degrees with distinction in Geodesy and Engineering Surveying from Wuhan University in 1997, 1999, and 2002, respectively. His main research interests include PPP, PPP-RTK, GNSS/INS integration technology and its applications.